\documentclass[runningheads]{llncs}
\usepackage[T1]{fontenc}

\usepackage{graphicx}
\usepackage{amsmath}
\usepackage{amssymb}

\usepackage{hyperref}

\usepackage[style=numeric, sorting=none]{biblatex}
\begin{document}
\title{Cryo-RL: automating prostate cancer cryoablation planning with reinforcement learning}
\titlerunning{Cryo-RL: cryoablation planning with reinforcement learning}

\author{Trixia Simangan \inst{1} \and
Ahmed Nadeem Abbasi \inst{2} \and
Yipeng Hu \inst{1} \and\\
Shaheer U. Saeed \inst{1,3,*}}
\authorrunning{Simangan et al.}
\institute{UCL Hawkes Institute \& Department of Medical Physics and Biomedical Engineering, University College London, United Kingdom \and
Aga Khan University Hospital, Pakistan \and
Centre for Bioengineering \& School of Engineering and Materials Science \& Digital Environment Research Institute, Queen Mary University of London, United Kingdom \\
*\email{shaheer.saeed@qmul.ac.uk}
}
\maketitle
\begin{abstract}
Cryoablation is a minimally invasive localised treatment for prostate cancer that destroys malignant tissue during de-freezing, while sparing surrounding healthy structures. Its success depends on accurate preoperative planning of cryoprobe placements to fully cover the tumour and avoid critical anatomy. This planning is currently manual, expertise-dependent, and time-consuming, leading to variability in treatment quality and limited scalability. 
In this work, we introduce Cryo-RL, a reinforcement learning framework that models cryoablation planning as a Markov decision process and learns an optimal policy for cryoprobe placement. Within a simulated environment that models clinical constraints and stochastic intraoperative variability, an agent sequentially selects cryoprobe positions and ice sphere diameters. Guided by a reward function based on tumour coverage, this agent learns a cryoablation strategy that leads to optimal cryoprobe placements without the need for any manually-designed plans.
Evaluated on 583 retrospective prostate cancer cases, Cryo-RL achieved over 8 percentage-point Dice improvements compared with the best automated baselines, based on geometric optimisation, and matched human expert performance while requiring substantially less planning time. These results highlight the potential of reinforcement learning to deliver clinically viable, reproducible, and efficient cryoablation plans.

Open-source implementation: \url{github.com/Trixsim23/Cryo-RL}
\end{abstract}

\section{Introduction}

Prostate cancer remains one of the most prevalent malignancies among men worldwide \cite{rawla2019epidemiology}, driving continued interest in minimally-invasive focal therapies that treat localised areas. These approaches are preferred over whole-gland therapies such as prostatectomy because they reduce side effects and better preserve healthy tissue \cite{faiella2024focal, wake2020mri}. Among them, cryoablation has emerged as a promising focal therapy, offering targeted destruction of cancerous tissue while minimising damage to surrounding healthy structures \cite{karwacki2025current}. The procedure involves inserting cryoprobes into the prostate gland, which then freeze spherical areas to cytotoxic temperatures below zero, such that the de-freezing process destroys the cells. 

The success of cryoablation critically depends on precise preoperative planning, typically performed using magnetic resonance (MR) images \cite{karwacki2025current, ramalingam2023image}. Clinicians must decide the number, position, and size of cryoprobe ice spheres to ensure complete tumour coverage while avoiding damage to surrounding healthy tissue \cite{ramalingam2023image} \footnote{While cryoprobe ice balls are usually ellipsoid-shaped, with the operator having some control over the shape and ellipsoid curvature, we model the ice balls as spherical for simplicity and to reduce the number of parameters that are to be optimised.}. 
This task is highly complex, requiring careful consideration of patient-specific anatomy and the spatial relationship between the tumour and nearby critical structures such as the urethra, rectum, and neurovascular bundles \cite{karwacki2025current, ramalingam2023image}. Despite its complexity, effective planning is considered essential for achieving therapeutic efficacy and minimising potential complications. 

Current preoperative planning for cryoablation is predominantly manual and guided by clinician heuristics, making it time-consuming, inconsistent across operators, and difficult to scale \cite{ramalingam2023image}. It also demands years of specialised training and is often restricted to high-volume expert centres. Variability in planning can lead to suboptimal cryoprobe placement, incomplete tumour coverage, or unintended damage to healthy tissue. These limitations underscore the pressing need for automated, data-driven planning approaches that can improve consistency, efficiency, and clinical outcomes across a broader range of healthcare settings.

While automated tools for cancer detection and segmentation have become increasingly prevalent, treatment planning remains comparatively under-explored and poses an ongoing challenge \cite{saha2024artificial, ozer2010supervised, pocius2024weakly, karam2025promptable, li2023prototypical}. Segmentation algorithms can accurately delineate tumour boundaries from medical images; however, translating these boundaries into actionable cryoablation plans is non-trivial due to the irregular and patient-specific nature of tumour morphology as well as the need to account for the energy delivery and its interaction with tissues \cite{lovegrove2018prostate}. Some prior studies have proposed optimisation methods that rely on simplified mathematical models of cryoprobe ice sphere formation \cite{lyons2019ablation, boas2017development, li2025iceball, chaitanya2022automatic}. These approaches incorporate some clinical constraints, however, there is a need for more adaptable and intelligent planning frameworks that can operate under realistic clinical conditions by accounting for factors such as intraoperative variability, limited probe insertions, and intermittent or staged ablations across multiple imaging sessions.

In this work, we formulate the cryoablation procedure as a Markov decision process (MDP), in which a neural network agent sequentially selects the placement and radius of cryoprobes. Each action results in a planned probe insertion and ice sphere formation, with stochastic noise introduced to model intraoperative variability. The agent is trained using the proximal policy optimisation reinforcement learning (RL) algorithm \cite{schulman2017proximal, saeed2024competing, tan2025spars}, guided by a reward function that encourages maximal overlap between the ice sphere and tumour, while penalising overlap with surrounding healthy tissue. By modelling the procedure in this way, the agent learns a planning policy that ablates cancerous tissue while preserving healthy structures, without relying on manually designed plans. Crucially, due to the explicit modelling for clinical constraints within our framework, the learned policy is robust factors such as a limited number of cryoprobe insertions, anatomical variability across patients and intraoperative variability.

The contributions of this work are summarised: 1) we propose a simulated cryoablation environment modelled as an MDP, incorporating realistic clinical constraints and intraoperative variability; 2) we design a reward function based on ice sphere–tumour overlap and healthy tissue avoidance, enabling effective reinforcement learning of cryoprobe placement strategies; 3) We evaluate our approach on a large retrospective dataset of 583 prostate cancer patients, demonstrating superior performance compared to heuristic and manual planning methods; 4) we release our codebase publicly to support reproducibility and further research in automated interventional planning.

\section{Methods}

\begin{figure}[!ht]
    \centering
    \includegraphics[width=0.7\textwidth]{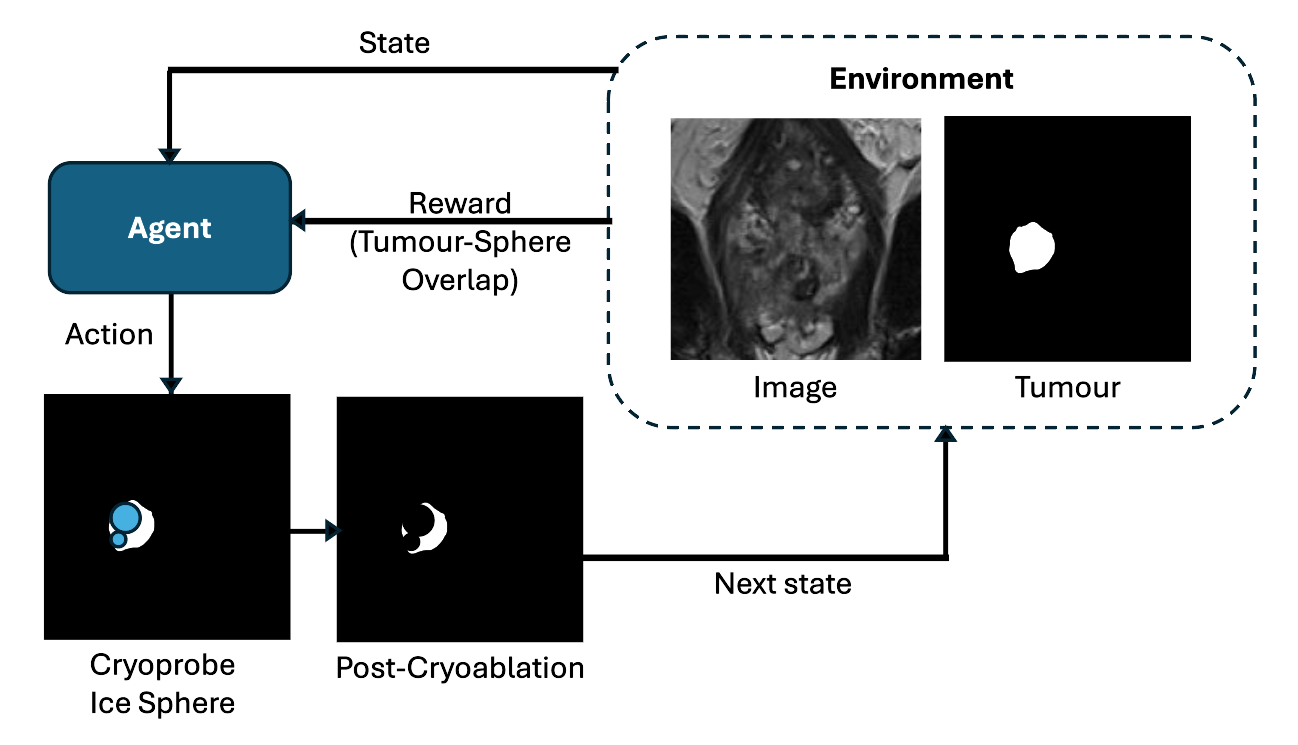}
    \caption{An overview of the Cryo-RL framework.}
    \label{fig:cryorl}
\end{figure}

An MDP environment simulates cryoablation with clinical constraints and intraoperative variability. The agent acts within this environment to decide cryoprobe placements which are optimised using RL, maximising reward based on ice sphere-tumour overlap. The framework is summarised in Fig. \ref{fig:cryorl}.

\subsection{Cryoablation environment}

\subsubsection{Pre-operative information for planning} % image and mask

The pre-operative MR image is denoted as $x\in\mathcal{X}$, where $\mathcal{X}$ is the domain of MR images. The corresponding tumour boundary segmentation mask is denoted as $y\in\mathcal{Y}$, where $\mathcal{Y}$ is the domain of tumour segmentation masks. These datapoints are available prior to the cryoablation procedure.

\subsubsection{Cryoablation Markov decision process environment} % input and output (state and action) and transitions

The pre-operative information may be considered part of an MDP environment. An agent observes a state of the environment and produces an action which influences the environment, generating a reward describing the efficacy of the action. The MDP environment may be defined with $(\mathcal{S}, \mathcal{A}, p, r, \pi)$, with each term further described below.

\textbf{States:} The observed state $s_t\in \mathcal{S}$ consists of the MR image $x$ and its tumour segmentation mask $y$, i.e., $s_t = \{x_t, y_t\}$, where $\mathcal{S}$ is the state space. The subscript $t$ denotes the time-step which corresponds to a single RL iteration corresponding to a single cryoablation visit and subsequent visits lead to an iteration of $t$.

\textbf{Actions:} The action $a_t \in\mathcal{A}$ is modelled as $a_t = \{i_{c,t}, j_{c,t}, k_{c,t}, d_{c,t}\}_{c=1}^{C}$ where $i,j,k,d\in\mathbb{R}_{\geq0}$, where $\mathcal{A}$ is the action space, $(i, j, k)$ denote the coordinates for the cryoprobe placement, $d$ denotes the diameter of the ice sphere and $C$ denotes the number of cryoprobe insertions. In practice, noise is added to these actions to model intraoperative variability, the scales of noise are fixed using reported intraoperative variability in the cryoablation procedure of 2.5mm in the three spatial dimensions and 5mm in the ice sphere diameter \cite{ramalingam2023image, wake2020mri, pushkarev2022study, karwacki2025current}. This noise is modelled using a truncated Gaussian distribution. The actions are modelled probabilistically, which ensures that the commonly-used treatment margins are accounted for within the modelling.  

\textbf{State transitions:} The action $a_t$ influences the observed state $s_t$ by changing the segmentation mask $x_t$ to remove the cryoprobe ice sphere, and the making corresponding changes to image $y_t$ (further described in Sec. \ref{sec:exp_data}). The next state is denoted as $s_{t+1} = \{x_{t+1}, y_{t+1}\}$. The state transition distribution conditioned on the state-action pair is given by $p: \mathcal{S} \times \mathcal{S} \times \mathcal{A} \rightarrow \left[0, 1\right]$, where assuming the next state $s_{t+1}$ as input, given the current state $s_t$ and action $a_t$, the probability of the next state is denoted as $p(s_{t+1}| s_t, a_t)$.

\textbf{Rewards:} The reward function $r: \mathcal{S} \times \mathcal{A} \rightarrow \mathbb{R}$, produces a reward $R_t = r(s_t, a_t)$, given the state-action pair as input. This reward measures the number of tumour voxels from the segmentation mask that fall within the ice sphere:

\begin{equation}
    R_{t} = \sum_{||(i,j,k)-(i_{c,t},j_{c,t},k_{c,t})||_2 ~\leq~ d_{c,t}/2} y_{i,j,k}
\end{equation}

where $y_{i,j,k}\in\{0,1\}$ is the binary tumour segmentation mask, with $y_{i,j,k}=1$ indicating that voxel $(i,j,k)$ belongs to the tumour tissue. Reward shaping is used to avoid repeated cryoprobe placements, as described in the open-source implementation. However, this has limited impact on performance, given sufficient training time as demonstrated in the ablation studies.

\textbf{Policy:} The policy $\pi:\mathcal{S}\times\mathcal{A}\rightarrow[0,1]$ denotes the probability of performing an action $a_t$. As an example, the probability of performing action $a_t$ is given by $\pi(a_t|s_t)$. Actions can be sampled according to the policy $a_t \sim \pi(\cdot | s_t)$ and this is what we call the agent. We can follow the state transition distribution for sampling next states $s_{t+1} \sim p(\cdot |s_t, a_t)$, the policy for sampling actions $a_t \sim \pi(\cdot | s_t)$ and the reward function for generating rewards $R_t = r(s_t, a_t)$. This leads to the collection of trajectory of state-action-reward triplets for multiple time-steps $(s_1, a_1, R_1, \dots, s_T, a_T, R_T)$, until the final time-step $T$, which represents the number of repeated visits to fully ablate the tumour.

\subsection{Optimisation using reinforcement learning}

For our optimisation, the goal is to find an optimal policy. To this end, we can model the policy as a parametric neural network $\pi(\cdot; \theta)$, with parameters $\theta$. This policy predicts parameters of a distribution from which actions can be sampled $a_t\sim\pi(\cdot|s_t; \theta)$. When we follow a policy, we can compute its cumulative reward as follows:

\begin{equation}
    Q^{\pi(\cdot|s_t;\theta)}(s_t, a_t) = \sum_{k=0}^T \gamma^k R_{t+k}
\end{equation}
where $\gamma$ is a discount factor for future rewards.
The optimisation then finds the optimal policy parameters that maximise the cumulative reward (implemented in practice as gradient ascent using the Proximal Policy Optimisation algorithm \cite{schulman2017proximal}):

\begin{equation}
    \theta^* = \arg \max_\theta \mathbb{E} \left[  Q^{\pi(\cdot|s_t;\theta)}(s_t, a_t) \right]
\end{equation}

The optimal policy parameters $\theta^*$ can then be used to sample optimal actions that denote optimal cryoprobe placements $a_t^*= \{i^*_{c,t}, j^*_{c,t}, k^*_{c,t}, d^*_{c,t}\}_{c=1}^{C} ~\sim\pi(\cdot|s_t; \theta^*)$.

\section{Experiments}

\subsection{Dataset}\label{sec:exp_data}

The data used in this project was obtained from the publicly available PROMIS trial data \cite{bosaily2015promis, ahmed2017diagnostic}, with 3-channel pelvic MR images (T2 Weighted, Diffusion and Apparent Diffusion Contrast) from 583 real prostate cancer patients. The segmentation masks for the tumour are available as part of the dataset and were obtained from trained imaging researchers with over 5 years of experience with MR imaging. The data was split into development and holdout sets using the ratio 7:3. All reported results are averaged across the 175 patient cases in the holdout set.

To simulate cryoablation, the tumour masks are erased at the location of the ice sphere placement and the location of the ice sphere in the image has voxels replaced with an average intensity value across the prostate gland to simulate a removed tumour (which may have intensity values different from the average across the gland). 

\subsection{Networks and hyper-parameters}

The policy network follows a 3D convolutional neural network architecture, for computational efficiency. The network is trained on a single Nvidia A100 GPU with the training lasting for approximately 12 hours, each forward pass taking 52ms and each environment iteration taking 31ms. 

Hyper-parameters were configured using a grid search over clinically plausible ranges, where results are presented in the ablation studies. In particular, the number of cryoprobe insertions per visit $C$, and the number of visits $T$ were tuned using this methodology.

The learning rate was set as $3\times10^{-4}$, the batch size was set as 512 and $\gamma$ was set as 0.9, all configured empirically.

\subsection{Comparisons}

Our Cryo-RL approach was compared firstly with a random action baseline to establish a lower bound for performance, where a cryoprobe was randomly inserted into a coordinate within the tumour. We also compared with a human imaging researcher, with 3 years of imaging experience, to establish the performance for the most commonly used current method. The human annotations were done using our own designed planning tool, available via the open-source repository. Another comparison was with the current state-of-the-art automated tool of geometric optimisation. And lastly we compared with a heuristic for placement which was to target the centre of the tumour. These comparisons provide a comprehensive overview of current approaches in practice.

The performance is evaluated in terms of the Dice score between the tumour segmentation mask and the placement of the cryoprobe ice sphere, and the standard deviation over patient cases.

% human
% geometric opt
% random
% center target

\subsection{Ablation studies} % removing reward shaping two components, number of needle palcements C , number of revisits allowed T

Our ablation studies focus on the components that impacted performance. The first is the addition of reward shaping to discourage repeated cryoprobe placements, where we quantify performance and training time until convergence. The other studies explore the impact of hyper-parameters for the number of cryoprobe placements per visit $C$ and the number of visits per patient $T$.

\section{Results}

\subsection{Comparisons}

Table \ref{tab:comparisons} shows that Cryo-RL was able to outperform all automated methods with substantial margins of over 8\% Dice percentage points. These differences were statistically significant (paired t-test) with a p-values in the range 0.002 - 0.031. It is also noteworthy that Cryo-RL outperformed the human expert, however, statistical significance was not observed for this comparison, with a p-value of 0.073. Furthermore, the planning time for Cryo-RL is substantially lower than the human and is comparable to the fastest automated planning method.

\begin{table}[!ht]
\centering
\begin{tabular}{|c|c|c|}
\hline
Planning Method & Performance / Dice & Planning Time / s\\
\hline
Cryo-RL (Ours) & 0.315 $\pm$ 0.098 & 2.4 \\
Human & 0.309 $\pm$ 0.194 & 538.6 \\
Geometric Optimisation & 0.235 $\pm$ 0.093 & 16.2\\
Centre Targetting & 0.156 $\pm$ 0.099 & 1.2 \\
Random Targetting & 0.132 $\pm$ 0.141 & 1.1 \\
\hline
\end{tabular}
\caption{Comparison of Cryo-RL with other planning approaches (times include data loading and offloading from GPU memory in environment, as in open-source code). }
\label{tab:comparisons}
\end{table}

\subsection{Ablation studies}

Table \ref{tab:ablations} shows that the reward shaping strategy which discourages repeated cryoprobe placements was effective in reducing the overall training time. However, the observed performance difference was not substantial, with statistical significance not observed for the comparison, with p-value 0.091.

Table \ref{tab:ablations} also shows that the optimal number of cryoprobes per visit is 5, where performance plateaus after this value. A statistically significant increase in performance is observed when the number of cryoprobes is increased from 4 to 5, with p-value 0.026. However, statistical significance is not observed for an increase from 5 to 6, with p-value 0.095. 

Lastly, Table \ref{tab:ablations} also summarises the impact of the number of revisits per patient, with a statistically significant increase observed between steps 3 and 4, with p-value 0.013. Beyond 4 revisits, however, performance plateaus with statistical significance not observed for the comparisons, with p-values 0.121 and 0.107.

These values for the number of cryoprobe insertions and revisits were tested due to their clinical plausibility, as reported in previous work \cite{ramalingam2023image, karwacki2025current, pushkarev2022study}.

\begin{table}[!ht]
\centering
\begin{tabular}{|c|c|c|c|}
\hline
Ablated Component & Value & Performance / Dice & Training Time / h\\
\hline
Reward Shaping & $\checkmark$ & 0.315 $\pm$ 0.098 & 11.7 \\
 & {\sffamily X} & 0.309 $\pm$ 0.104 & 25.4\\
\hline
Number of Cryoprobes $C$ & 1 & 0.139 $\pm$ 0.125 & - \\
 & 2 & 0.147 $\pm$ 0.155 & - \\
 & 3 & 0.154 $\pm$ 0.126 & -\\
 & 4 & 0.258 $\pm$ 0.102 & -\\
 & 5 & 0.315 $\pm$ 0.098 & -\\
 & 6 & 0.314 $\pm$ 0.113 & -\\
 \hline
Number of Revisits $T$ & 1 & 0.205 $\pm$ 0.126 & -\\
 & 2 & 0.274 $\pm$ 0.115 & -\\
 & 3 & 0.303 $\pm$ 0.096 & -\\
 & 4 & 0.315 $\pm$ 0.098 & -\\
 & 5 & 0.311 $\pm$ 0.087 & -\\
 & 6 & 0.312 $\pm$ 0.108 & -\\
\hline
\end{tabular}
\caption{Ablations and hyper-parameter variations for the Cryo-RL framework.}
\label{tab:ablations}
\end{table}

\subsection{Qualitative results}

\begin{figure}
    \centering
    \includegraphics[width=0.8\textwidth]{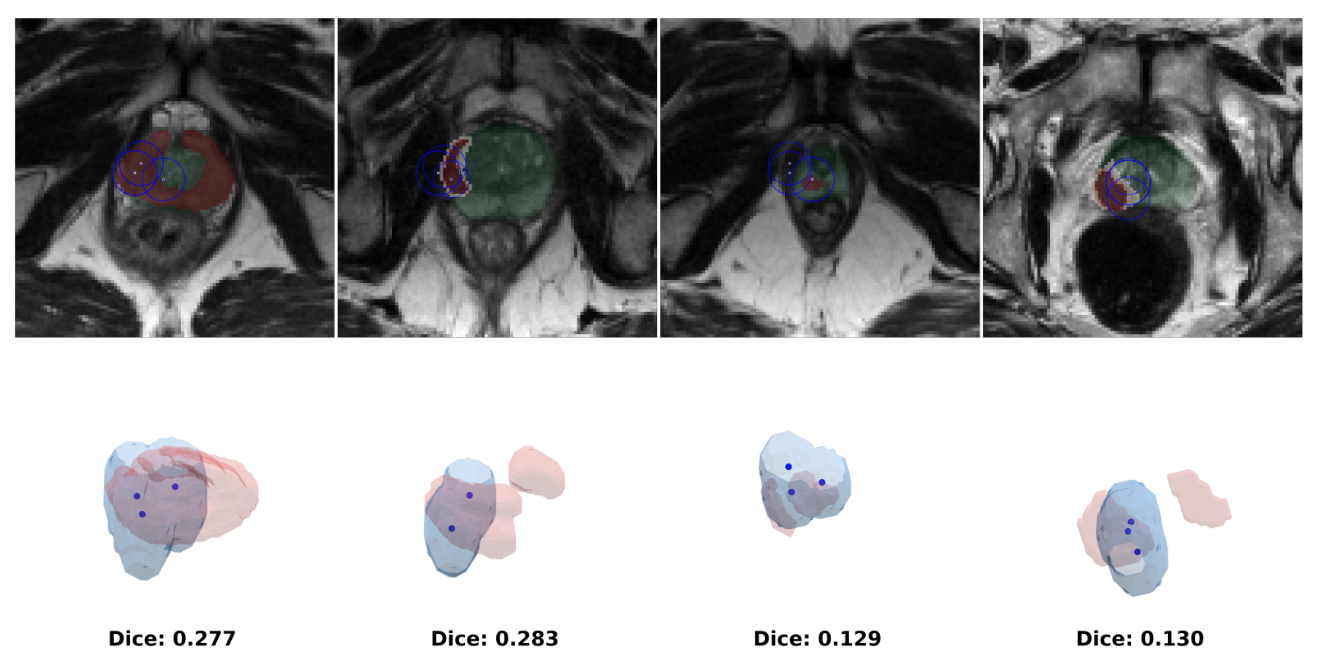}
    \caption{Cryo-RL ablation plans for first visit ($t=1$) showing prostate in green, lesion in red and ablation area in blue, with dots indicating the projection of the centre-points of the placed cryoprobes. The top shows 2D slices from the 3D volume overlaid onto the preoperative MR image slice and the bottom shows a 3D view.}
    \label{fig:grid_examples}
\end{figure}

Fig. \ref{fig:grid_examples} shows the examples of cryoprobe placements from Cryo-RL. Cryo-RL tends to over-ablate small tumours while underablating larger tumours or areas. The overall under-ablation is expected as there are multiple visits over which the tumour can be fully ablated (the examples in Fig. \ref{fig:grid_examples}, show results from the first visit $t=1$).

\section{Discussion}

The results demonstrate that Cryo-RL substantially outperforms existing automated planning methods for prostate cancer cryoablation, while maintaining planning time comparable to the fastest automated method. Notably, Cryo-RL also surpassed the performance of a human expert, although this difference did not reach statistical significance. This finding suggests that reinforcement learning can capture planning strategies on par with expert decision-making while providing substantial efficiency gains, potentially enabling more consistent treatment quality across centres with varying levels of expertise. Although a comparison with multiple expert clinicians would be necessary before deployment, the current comparisons with a human imaging expert demonstrate feasibility and provide a strong justification for future work expanding the framework.

Ablation studies provide further insight into the factors influencing Cryo-RL’s performance. Reward shaping to discourage repeated probe placement was effective in reducing training time by more than half, though its effect on final Dice score was modest. Analysis of the number of cryoprobes per visit and the number of revisits improved results up to a plateau at clinically plausible values. 

Qualitative results in Fig. \ref{fig:grid_examples} show clinically plausible plans for the first cryoablation visit (t=1) across different patient cases. Cryo-RL often targets a single area per visit — a behaviour that emerged naturally from RL training rather than manual design — likely because multiple smaller spheres can better match complex tumour boundaries. In small tumours, over-ablation was observed, potentially due to the intraoperative variability modelled as noise, prompting the agent to include surrounding tissue to maximise the chance of complete tumour ablation. While Cryo-RL’s plans were assessed both qualitatively and quantitatively here, a prospective evaluation may yield further insights. Moreover, the proposed framework could be extended to other energy sources for tumour ablation, such as high-intensity focused ultrasound or radio-frequency ablation.

\section{Conclusion}

We presented Cryo-RL, a reinforcement learning framework for automated preoperative planning of prostate cancer cryoablation. By modelling the task as an MDP and training an agent within a clinically constrained simulation, Cryo-RL learns planning policies that achieve substantial improvements over existing automated methods and perform comparably to human experts, while reducing planning time by orders of magnitude compared to human planners. The approach produces clinically plausible strategies, suggesting its potential to support more accessible, efficient, and standardised delivery of focal therapies.

\section*{Acknowledgements}

This work is supported by the International Alliance for Cancer Early Detection, an alliance between Cancer Research UK [EDDAPA-2024/100014] \& [C73666/A31378], Canary Center at Stanford University, the University of Cambridge, OHSU Knight Cancer Institute, University College London and the University of Manchester; and National Institute for Health Research University College London Hospitals Biomedical Research Centre.

\printbibliography

\end{document}